\documentclass[10pt,twocolumn,letterpaper]{article}

\usepackage{cvpr}              
\usepackage{makecell}
\usepackage{multirow}
\usepackage[accsupp]{axessibility} 
\usepackage[T1]{fontenc}
\usepackage{booktabs} 
\usepackage{url}
\usepackage{multirow}
\usepackage[table]{xcolor}
\usepackage{subcaption}  
\usepackage{algorithm}
\usepackage{algpseudocode}
\usepackage{tcolorbox}
\usepackage{soul}       
\usepackage{xcolor}     

\definecolor{lightyellow}{RGB}{255, 255, 180}
\definecolor{lightblue}{RGB}{180, 220, 255}
\definecolor{lightorange}{RGB}{255, 210, 150}
\definecolor{lightgreen}{RGB}{190, 255, 190}

\usepackage{graphicx}

\definecolor{cvprblue}{rgb}{0.21,0.49,0.74}

\usepackage[pagebackref,breaklinks,colorlinks,allcolors=cvprblue]{hyperref}

\usepackage{tcolorbox}
\tcbuselibrary{theorems}

\newtcolorbox[auto counter, number within=section]{promptbox}[2][]{
  colback=gray!5,
  colframe=gray!60,
  fonttitle=\bfseries,
  title={Prompt~\thetcbcounter: #2},
  #1
}
\def\paperID{40157} 
\def\confName{CVPR}
\def\confYear{2026}

\title{Beyond the Global Scores: Fine-Grained Token Grounding as a Robust Detector of LVLM Hallucinations
\vspace{-4mm}
}

\author{
Tuan Dung Nguyen$^{1,2}$ \and Minh Khoi Ho$^{1,4}$ \and Qi Chen$^{2}$ \and Yutong Xie$^{4}$ \and Cam-Tu Nguyen$^{3}$ \quad Minh Khoi Nguyen$^{1}$ \and Dang Huy Pham Nguyen$^{1,5}$ \quad Anton van den Hengel$^{2}$ \and Johan W. Verjans$^{2}$ \quad Phi Le Nguyen$^{1}$ \thanks{Corresponding authors} \quad 
Vu Minh Hieu Phan$^{2}$\footnotemark[1]
\\ 
$^{1}$ Hanoi University of Science and Technology \\
$^{2}$ Australian Institute for Machine Learning, University of Adelaide \\
$^{3}$ Nanjing University \quad
$^{4}$ Mohamed bin Zayed University of Artificial Intelligence \\
$^{5}$ Hanoi-Amsterdam High School for the Gifted 
\vspace{-4mm}
}
\begin{document}

\maketitle
\begin{abstract}

Large vision-language models (LVLMs) achieve strong performance on visual reasoning tasks but remain highly susceptible to hallucination. Existing detection methods predominantly rely on coarse, whole-image measures of how an object token relates to the input image. This global strategy is 
limited: hallucinated tokens may exhibit weak but widely scattered correlations across many local regions, which aggregate into deceptively high overall relevance, thus evading the current global hallucination detectors. 
We begin with a simple yet critical observation: \textup{a faithful object token must be strongly grounded in a specific image region.}
Building on this insight, we introduce a patch-level hallucination detection framework that examines fine-grained
token-level interactions across model layers.
Our analysis uncovers two characteristic signatures of hallucinated tokens:
\textbf{(i)} they yield diffuse, non-localized attention patterns, in contrast to the compact, well-focused attention 
seen in faithful tokens; and
\textbf{(ii)} they fail to exhibit meaningful semantic alignment with any visual region.
Guided by these findings, we develop a lightweight and interpretable detection method that leverages \textbf{patch-level} statistical features, combined with hidden-layer representations. Our approach achieves up to 90\% accuracy in token-level hallucination detection, demonstrating the superiority of fine-grained structural analysis for detecting hallucinations. The code is available  \href{https://github.com/tuandung2812alt3/token-grounding-detector.git}{here}. 
\end{abstract}
\vspace{-3mm}
\section{Introduction}

\begin{figure*}[htbp]
    \centering
\includegraphics[width=0.8\textwidth]{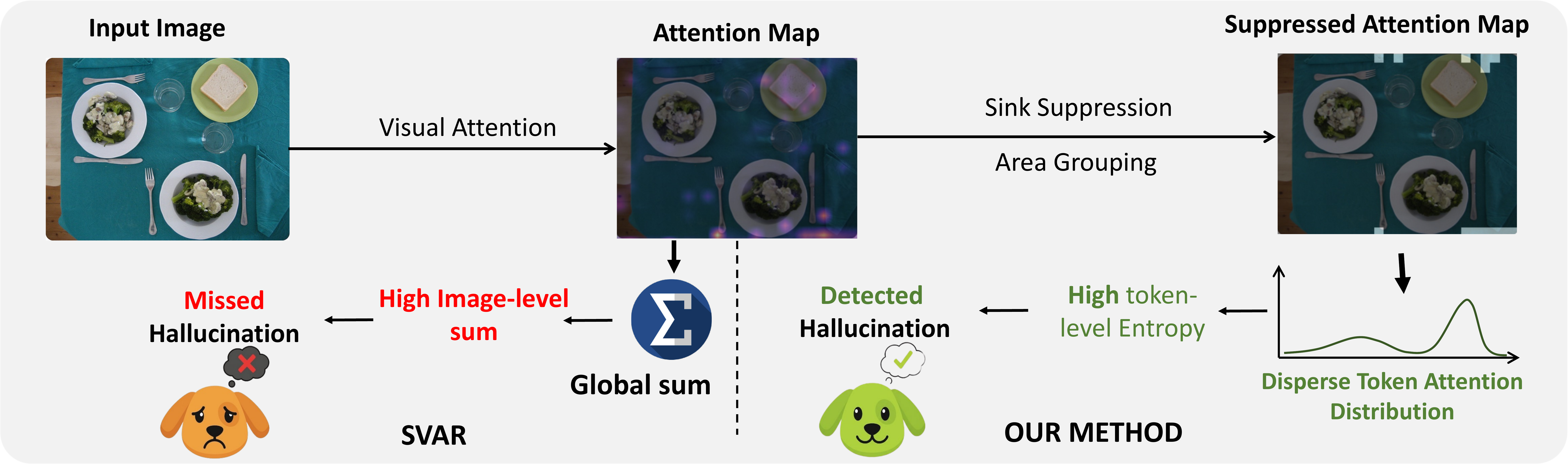} 
\caption{\textit{Left:} 
SVAR~\cite{jiang2025devils} captures \textbf{image-level statistics} by globally summing the attention across the image, ignoring local attention structures. This triggers a false alarm when the local noisy attention appears. \textit{Right:} Our proposed “attention dispersion score” encoded \textbf{token-level attention distribution}, quantifying how much the model allocates attention to various local regions simultaneously. Spreading the focus across multiple regions, \ie high entropy, indicates hallucination. Specifically, our pipeline first suppresses noisy regions and detects focused objects by grouping, and then quantifies the attention entropy of the token-level distributions. Low entropy means the model “sharply” focuses on particular objects, thus being likely to predict true objects. In other words, high entropy means the model is unfocused when predicting, indicating hallucination.
}  
\label{fig:compare}
\end{figure*}
\label{sec:introduction}
Large Vision-Language Models (LVLMs), including LLaVA~\cite{liu2023visual}, InstructBLIP~\cite{blip}, Otter~\cite{li2023otter}, mPLUG-Owl~\cite{ye2023mplugowl}, MiniGPT-4~\cite{zhu2023minigpt4}, and Qwen-VL~\cite{bai2023qwen}, have achieved impressive performance across a wide spectrum of multimodal tasks such as image captioning, visual question answering (VQA), and visual conversation systems. Despite this progress, these models remain vulnerable to a critical failure: visual hallucination, where they describe objects or attributes that do not appear in the image~\cite{rohrbach2018object,pope}. Such errors pose risks in high-stakes settings like robotics and healthcare, making accurate detection of hallucinated objects essential for improving LVLM trustworthiness.

Most existing hallucination-detection methods follow a 
\textit{global-statistics}
paradigm: they assess how strongly the object token in the output relates to the entire image. Prior works, for example, aggregate total attention across all patches~\cite{jiang2025devils}, analyze attention from preceding tokens~\cite{fieback2024metatoken}, examine output token probabilities~\cite{fieback2024metatoken}, or compute embedding similarities using a separate pretrained model~\cite{park2025halloc}. In essence, current approaches evaluate the global image-token correlation rather than examining grounding at the level of specific image regions. This coarse-grained strategy introduces a fundamental limitation. A hallucinated token (i.e., one that is not grounded in any actual image region) may still exhibit weak but uniformly distributed relevance across many patches (Fig.~\ref{fig:compare}, Left). When these diffuse signals are aggregated, the object token appears highly correlated with the whole image, leading existing methods to mistakenly categorize hallucinations as truthful.

From this observation, we derive a key insight: \textit{if an object truly exists in the image, its corresponding token must align strongly with a specific region where that object is visually present.} Consequently, hallucination detection must move beyond \textit{whole-image} analysis and instead evaluate how well a token aligns with individual \textit{patches}. Our experiments validate this intuition. We find that hallucinated tokens consistently show low attention 
with every individual region, while simultaneously producing weak, scattershot correlations across many unrelated areas. Fig.~\ref{fig:compare} highlights this discrepancy: the state-of-the-art detector SVAR~\cite{jiang2025devils}, using global statistics, frequently misclassifies hallucinations due to the attention sink phenomenon, when models place concentrated attention on irrelevant regions ( Fig.~\ref{fig:feature-similarity}).

\begin{figure*}[htbp]
    \centering
\includegraphics[width=1.0\textwidth]{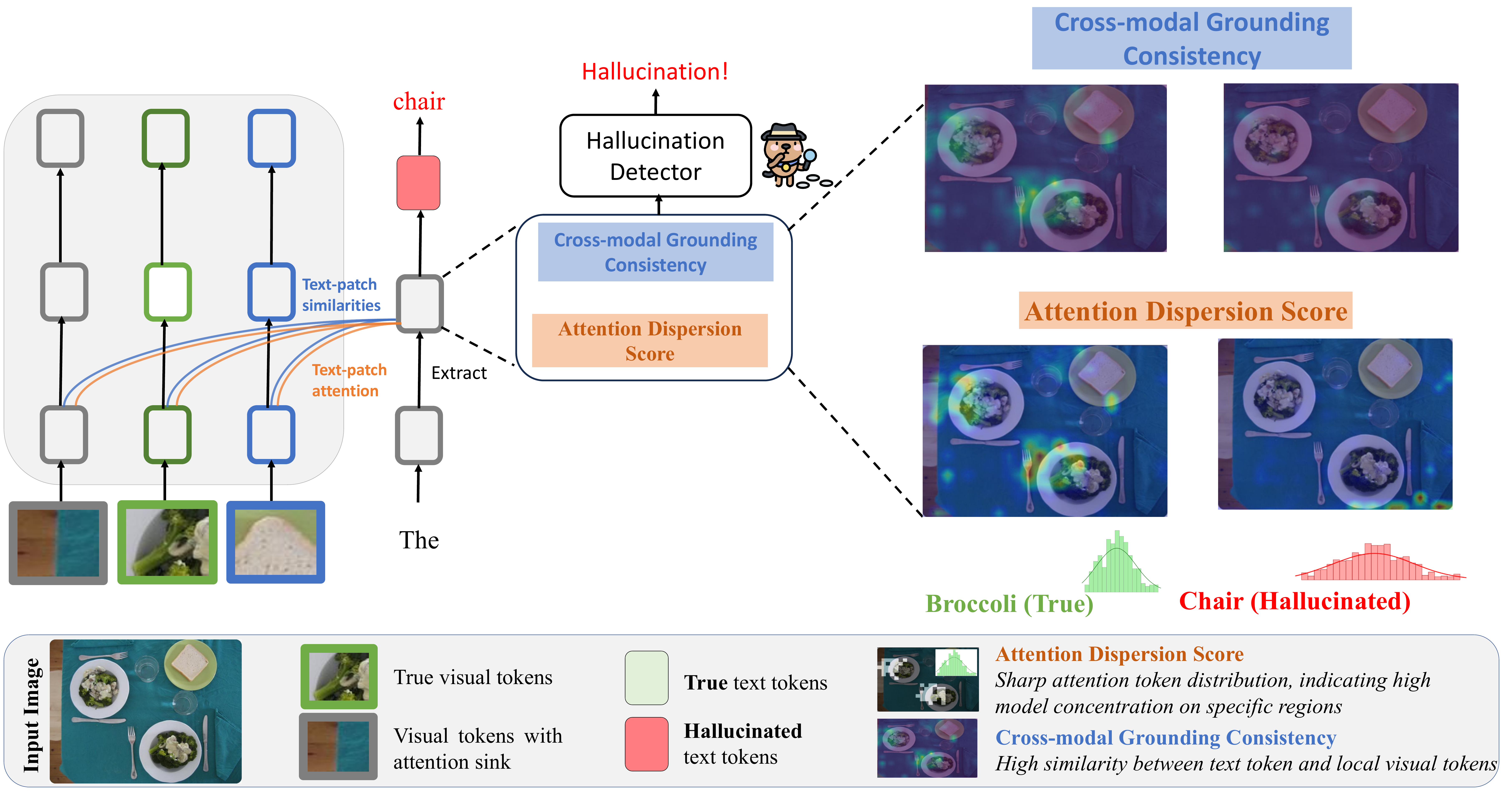} 
\caption{\textbf{Overview of our token-level hallucination detection framework}. We reveal two key indicators of hallucination:
\textbf{(1)} \textit{Attention Dispersion Score (ADS)}, hallucinated tokens exhibit highly diffused, non-localized attention across local image patches, while faithful tokens show concentrated focus; and
\textbf{(2)} \textit{Cross-modal Grounding Consistency (CGC)}, hallucinated text tokens exhibit low alignment with any object regions, as shown in the right figure.
Our proposed metrics encode \textit{local structures} of the VLMs' behaviors, enabling lightweight and explainable hallucination detection robust to attention sink scenarios. }  
\label{fig:intro_method}
\end{figure*}

These observations expose a critical gap in current methods and underscore the need for fine-grained \textit{patch-level} grounding analysis for reliable hallucination detection. Our approach addresses this challenge by explicitly analyzing the spatial structure of cross-modal attention maps and measuring how accurately tokens align with the specific regions in which the corresponding objects would appear.
Through patch-level analysis, we identify distinct internal signatures of hallucinated tokens:
\textbf{(1)} they display diffuse and scattered attention across patches, and
\textbf{(2)} they exhibit 
low semantic similarity to all image regions.
These behaviors are illustrated in Figures~\ref{fig:entropy_quant} and~\ref{fig:feature-similarity}.
From these findings, we introduce two new metrics \textit{Attention dispersion score} and \textit{Cross-modal grounding consistency}, shown in Fig.~\ref{fig:intro_method}, which enable accurate discrimination between hallucinated and grounded tokens. Finally, we demonstrate that these patch-level behavior metrics lead to state-of-the-art hallucination detection performance, surpassing existing methods. Our contributions are as follows:
\begin{itemize}
    \item We identify that analyzing textual tokens' relationships with specific image regions at a patch-level granularity enables effective hallucination detection. Building on this insight, we introduce new techniques to reveal the \textit{structural behaviors} that emerge within LVLM hidden layers during hallucination, leveraging fine-grained attention distributions and internal cross-modal feature interactions.
    \item To capture these structural behaviors, we propose two novel metrics: \textit{Attention Dispersion Score} (ADS) and \textit{Cross-modal Grounding Consistency} (CGC), which measure how well a token is visually grounded in the image. These metrics enable systematic analysis of token-level hallucination patterns.
    \item We present two key findings: (i) Real object tokens exhibit compact, well-localized attention on the relevant visual patches, while hallucinated tokens display diffuse, scattered attention across unrelated regions; (ii) Genuine tokens maintain high semantic similarity with the corresponding image patches, whereas hallucinated tokens show weak or inconsistent visual-semantic alignment. Together, these findings suggest that hallucinations arise primarily from overreliance on language priors, rather than deficiencies in the visual encoder.
    \item Using these insights, we develop a simple yet highly effective token-level hallucination detector that integrates hidden-layer representations with statistical measures such as entropy, patch similarity, and confidence alignment. Our framework achieves state-of-the-art performance, reaching up to 90\% detection accuracy,  the strength of our methodological contributions.
\end{itemize}

\section{Related Works}
\label{sec:related_works}
\textbf{Large Vision-Language Models}
Large Vision-Language Models (LVLMs) are highly capable multimodal models that
integrate high-capacity visual encoders with instruction-tuned language models to enable unified multimodal reasoning and generation. Foundational models such as LLaVA~\cite{liu2023visual}, InstructBLIP~\cite{blip}, Flamingo~\cite{flamingo}, MiniGPT-4~\cite{zhu2023minigpt4}, InternVL \cite{internvl}, Qwen-VL~\cite{bai2023qwen},  Gemma~\cite{gemma} are state-of-the-art representatives, with increasingly impressive benchmark results on diverse visual reasoning tasks. Despite remarkable progress, LVLMs remain prone to \textit{object hallucinations}—where the model describes objects that do not exist in the image~\cite{pope, rohrbach2018object}. These hallucination failures pose key challenges for improving LVLM visual reliability and trustworthiness. \\
\textbf{Object Hallucination Benchmarks}. 
To quantify hallucination in multimodal generation, several evaluation benchmarks have emerged. Early efforts such as CHAIR~\cite{rohrbach2018object} derive metrics for object hallucination in open-ended captioning, while POPE~\cite{pope} provides a rigorous benchmark revealing strong links between hallucination frequency and language co-occurrence priors. ROPE~\cite{rope} extends these settings to multi-object scenes. He~\cite{he2025evaluating} examine hallucination under object-deletion perturbations, showing that LVLMs still struggle with language bias.  
Unlike prior work that focuses primarily on closed-form or multiple-choice settings. \\
\textbf{Object Hallucination Detection}
Hallucination detection in Vision Language Models techniques span a broad methodological spectrum. Some approaches consist of  
Logic-based consistency checking~\cite{manakul-etal-2023-selfcheckgpt, wu2024logical} which verifies cross-question coherence; visual grounding verification~\cite{woodpecker} which uses external vision models to confirm object mentions. Sentence-level detectors~\cite{xiao2025detecting, gunjal2024detecting} train classifiers on labelled hallucinated outputs, leveraging features such as object mention patterns, improbability scores, and external retrieval signals. Utilizing deeper LVLM features, DHCP \cite{dhcp} extracts raw cross-attention weights and feed the attention vectors to light-weight classifiers to classify sentence's hallucination.
\cite{pixel_to_tokens} enables the LVLM to better localize objects during inference, and help mitigate hallucination, through introducing \textit{virtual tokens} and utilizing a Mask R-CNN decoder during finetuning.
While achieving decent results for hallucination detection, the mentioned methods fails to localize the exact hallucinated tokens within the sentence. In contrast, recent efforst ~\cite{fieback2024metatoken, park2025halloc, jiang2025devils} have targeted hallucination at the token-level. MetaToken \cite{fieback2024metatoken} emphasizes on feature engineering, extracting different handcrafted features from the LVLMs' outputs during decoding, such as tokens' probabilistic entropy, attention distribution... and uses traditional Machine Learning models for detection. SVAR also follows a similar approach , with the features being the ratio of attention allocated to the image tokens by the language decoders' middle layers. HalLoc~\cite{park2025halloc} utilizes a Bert-based multimodal embedding model \cite{visualbert} and four classification heads are employed to classify whether each token is hallucinatory. However, these methods do not take into account \textit{patch-level grounding analysis}, as in how the language tokens align with different visual areas of the image. Some works \cite{dhcp, jiang2025devils}, while utilize attention weights for detecting hallucination, only use global statistics such as raw cross-attention weights \cite{dhcp} or summed ratio of visual attention \cite{jiang2025devils}, without considering the spatial inner structure and distribution of the visual attention map.

\vspace{-3mm}
\begin{figure*}[t]
  \centering
  \includegraphics[width=0.89\textwidth]{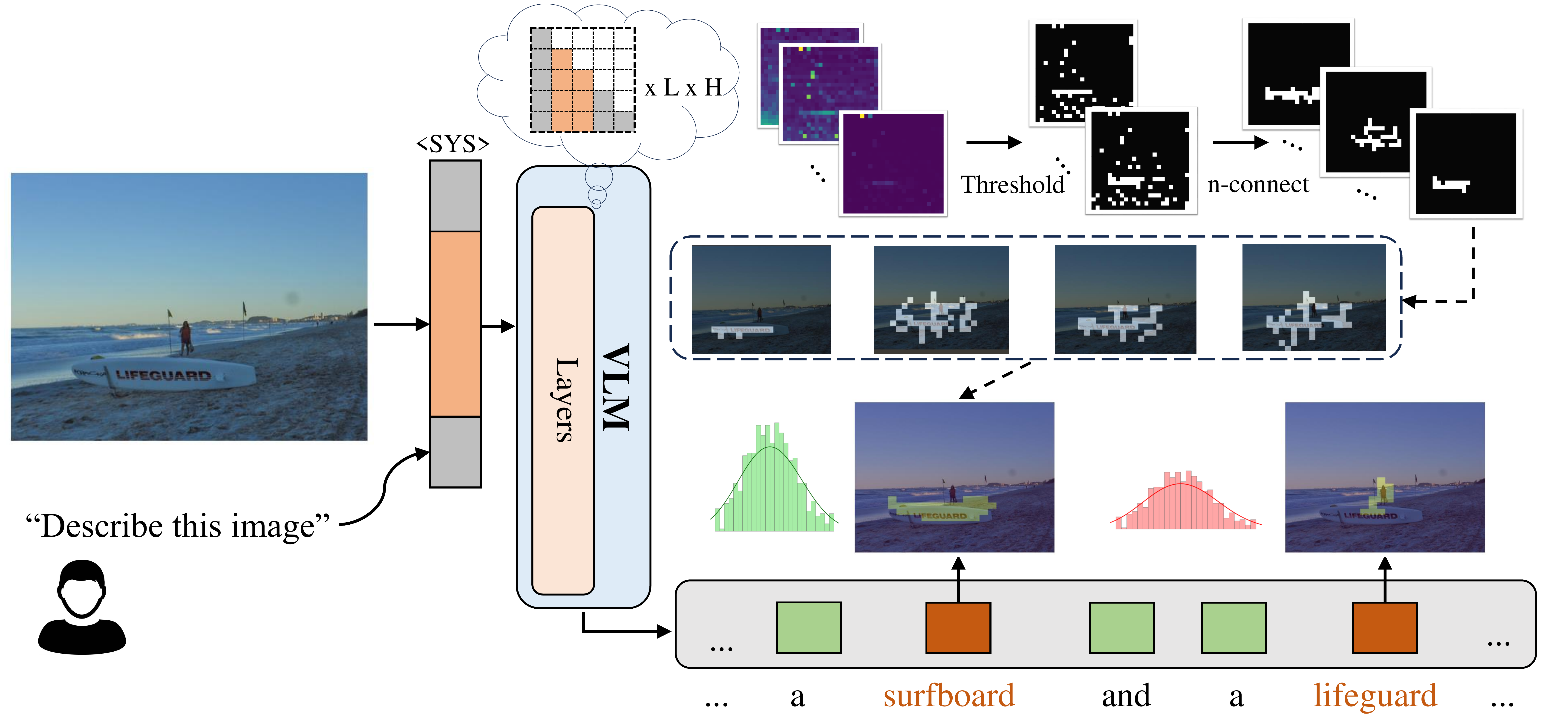}
  \caption{Illustration of the \textbf{Attention Dispersion Score} (ADS) computation. After predicting an object token, we extract the text-to-patch cross-modal attention map. The top $k\%$ activations are kept to isolate highly focused regions. Then, we form the attended object regions and suppress attention sinks by applying $N$-connected component. Finally, the proposed ADS score is computed by quantifying the entropy of the resulting token-level attention distribution. The \textcolor[HTML]{B1D095}{peaky} distribution indicates sharp object focus, indicating \textcolor[HTML]{B1D095}{real} object token predictions. In contrast, the \textcolor[HTML]{B86029}{uniform} patch-wise distribution implies that the model scatters its attention when predicting, indicating \textcolor[HTML]{B86029}{hallucinated} predictions.
}
  \label{fig:entropy_layer}
\end{figure*}

\section{Token-level Analysis of Large Vision-Language Models}
\label{sec:token_level_analysis}
We conduct token-level investigation, revealing the distinct interactions between hallucinated tokens and the image patch tokens, thus, useful to detect hallucination.
Our study yields two key findings: \textit{hallucinated tokens (i) have diffused attention, and (ii) show low grounding with any image patches.} To quantify the two behaviors, we introduce two \textbf{patch-level structural statistics}: (i) Attention Dispersion Score, and (ii) Cross-modal Grounding Consistency. We also find a minor finding: \textit{our proposed statistics has positive correlations with the model confidence.} 

\subsection{Setup}
We report results on a range of popular open-source LVLMs: LLaVA-V1.5-7B, Qwen2.5-VL-7B and InternVL-2.5-8B. Models are prompted with \textit{``Describe this image.''} and decoded greedily unless otherwise stated. Following the standard procedures in ~\cite{leng2024vcd, jiang2025devils}, we run experiments on a subset of 4000 MS-COCO-2014~\cite{lin2014microsoft} images from the validation set, using a 90/10 split, and classify generated object tokens as either True Object or Hallucinated Object using CHAIR~\cite{rohrbach2018object} but using GPT-4o for an extra layer of semantic verification.
\subsection{Attention Disperse Score}
\label{sec:3.1_attention_entropy}

\begin{figure}[t!]
  \centering
  \includegraphics[width=1.0\linewidth]{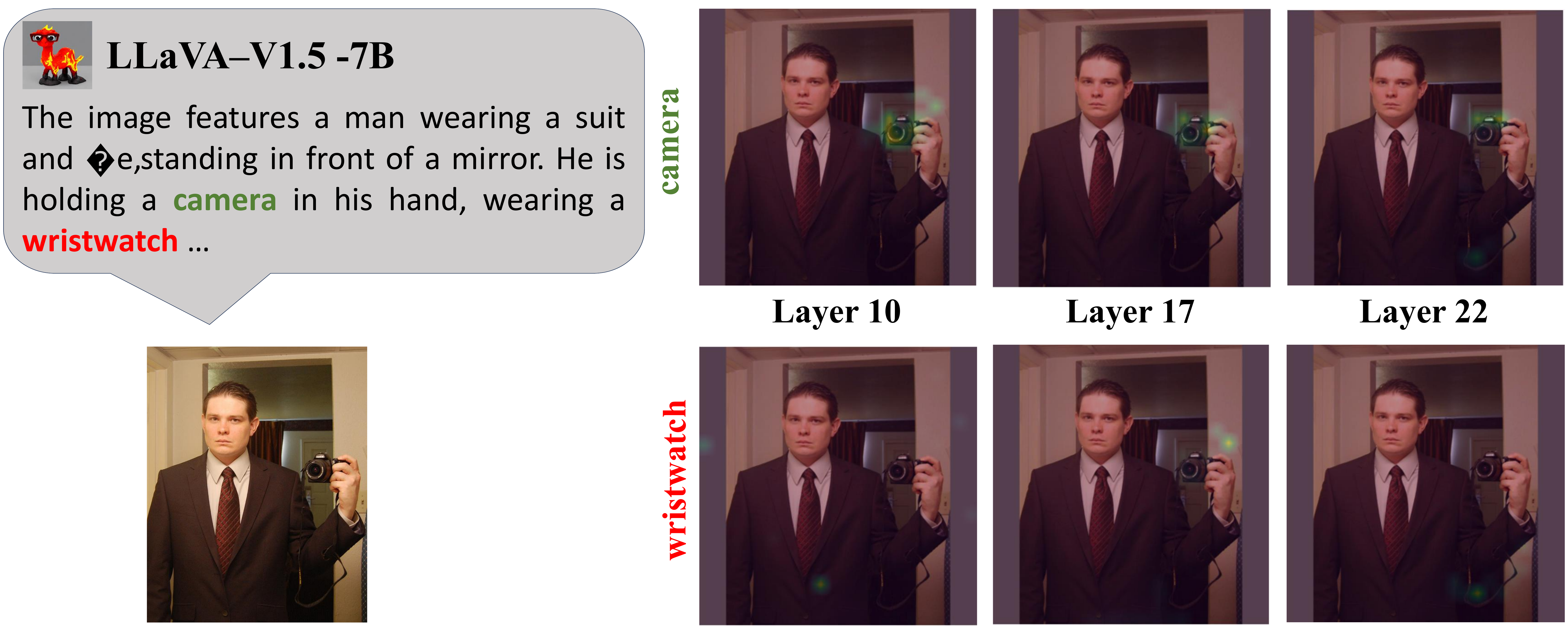}
  \caption{The visual attention maps for or a true token (``camera'', top) vs.\ a hallucinated token (``wristwatch'', bottom) across layers (10, 17, 22). True tokens reveal a focused attention pattern aligned with the object’s location, while hallucination tokens have scattered attention across the image.}
  \label{fig:entropy_quant}
\end{figure}

\begin{figure*}[t]
  \centering
\includegraphics[width=0.98\linewidth]{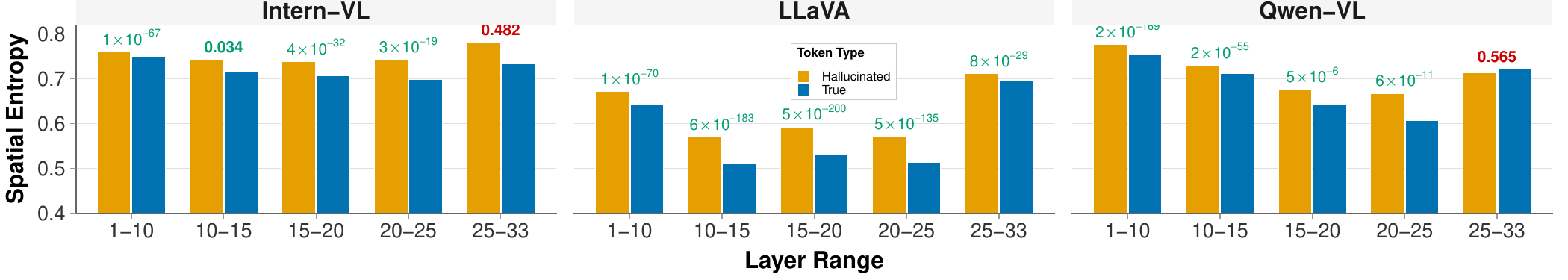}
  \caption{Layer-wise attention entropy of \emph{true} vs.\ \emph{hallucinated} tokens across LVLMs (lower is better; more focused). Reported $p$-values indicate strong separation in early/mid layers.}
  \label{fig:attn_layer_bars}
\end{figure*} 

\paragraph{Discovery.}
LVLMs exhibits more \emph{scattered} attention distributions over patches when producing hallucinated tokens, whereas correct tokens 
display \emph{compact} and \emph{localized} attention.  
Intuitively, we can look at the following qualitative example in Fig
\ref{fig:entropy_quant}, where  the model is able to allocate enhanced attention to a focused area if the object is not hallucinatory, while failing to do so in hallucinated objects. \\
\textbf{Spatial entropy reveals hallucinated tokens.}
To validate the hypothesis, we propose attention entropy to model the distribution of the attention map, and quantify the \textit{compactness} of attention distribution. For a generated token $t$ at layer $n$, let $\mathbf{A}^{(n,h)}_{t} \in \mathbb{R}^{|\mathcal{P}|}$ denote the attention from head $h$ over the patch set $\mathcal{P}$ (e.g., a $24 \times 24$ patch grid).
We first average across $H$ heads to obtain a layer-wise attention map over visual patches:
\vspace{-3mm}

\begin{equation}
\bar{\mathbf{A}}^{(n)}_{t} = \frac{1}{H}\sum_{h=1}^H \mathbf{A}^{(n,h)}_{t}.
\end{equation}

To isolate the most salient activations, we retain only the top-$x\%$ of patch responses (empirically $x=10$), defining a \textit{foreground} set $\mathcal{F}^{(n)}_t$ and its complement $\mathcal{B}^{(n)}_t = \mathcal{P} \setminus \mathcal{F}^{(n)}_t$.
The retained patches are then grouped into 8-connected components $\mathcal{C}^{(n)}_{t}$, and small spurious blobs (area $< \tau_{ADS}$) are suppressed to filter out attention sinks, yielding valid components $\mathcal{C}^{(n)*}_{t}$.

\begin{figure}[tbp]
\centering
\includegraphics[width=0.65\linewidth]{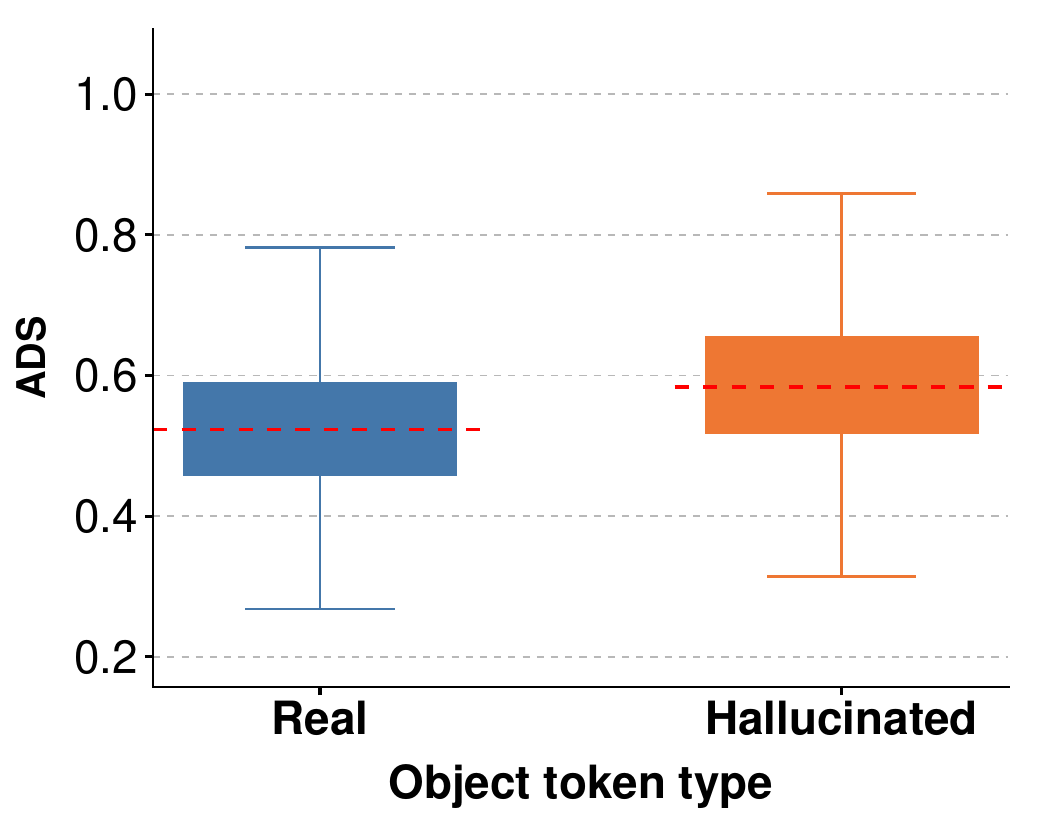}
\caption{Attention dispersion distribution for \emph{true} vs.\ \emph{hallucinated} tokens for LLaVA-1.5-7B.}
\label{fig:box_entropy}
\end{figure}

We decompose the spatial \textit{compactness} of attention into two complementary signals:
(i) the \textit{foreground blob mass}
\begin{equation}
m^{(n)}_{t} = \sum_{c \in \mathcal{C}^{(n)*}_{t}} \sum_{p \in c} \bar{\mathbf{A}}^{(n)}_{t}(p),
\label{eq:blob}
\end{equation}
measuring how much total attention the valid components $\mathcal{C}^{(n)*}_{t}$ capture, and
(ii) the \textit{background entropy}, which quantifies the entropy of residual attention outside the foreground. 
We renormalize over background patches to obtain a valid distribution:
\begin{equation}
\mathbf{E}^{(n)}_{t}(p) 
= \frac{\bar{\mathbf{A}}^{(n)}_{t}(p)}
{\sum_{p' \in \mathcal{B}^{(n)}_t} \bar{\mathbf{A}}^{(n)}_{t}(p')}, 
\quad p \in \mathcal{B}^{(n)}_t,
\end{equation}
and compute its normalized Shannon entropy:
\begin{equation}
\hat{H}^{(n)}_{t} 
= \frac{-\sum_{p \in \mathcal{B}^{(n)}_t} 
{\mathbf{E}}^{(n)}_{t}(p) 
\log {\mathbf{E}}^{(n)}_{t}(p)}
{\log |\mathcal{P}|}.
\end{equation}
Dividing by $\log |\mathcal{P}|$ maps the entropy to $[0,1]$, 
where $0$ indicates all residual mass concentrated on a single 
background patch and $1$ indicates a uniform spread across all patches.
 The per-layer Attention Dispersion Score combines both signals:
\begin{equation}
\mathrm{ADS}^{(n)}_{t} = (1 - m^{(n)}_{t}) \cdot {\hat{H}^{(n)}_{t}}.
\end{equation}
A low ADS score indicates that attention is concentrated in coherent foreground blobs with quiet background (compact, grounded focus), while high ADS score indicates scattered blob mass and noisy residual activation. We summarize the calculation of our ADS metric in Fig.~\ref{fig:entropy_layer}. \\
\textbf{Remarks.} As shown in Fig.~\ref{fig:attn_layer_bars}, grounded tokens consistently display lower ADS than hallucinations in early and mid layers, reflecting stronger visual focus before linguistic priors dominate at depth. The mid-layer averaging further stabilizes this separation across LVLMs. Figure \ref{fig:box_entropy} reports the distribution of ADS in middle layers (layer 10-25), which shows a clear separation in attention dispersion between True and Hallucinated tokens. \\
\textbf{Attention Dispersion Score (ADS) as a hallucination detector.}
To quickly validate the hallucination discriminativeness of the Attention Dispersion Score, we construct a simple linear classifier trained on the per-layer ADS vector. From Table~\ref{tab:mlp_entropy} displays that F1 scores lie in $0.73$–$0.77$, establishing \emph{attention compactness} as a useful signal to detect hallucinated tokens. \\
\textbf{Discussion.}
Recent studies have explored the use of attention entropy for \textit{object localization}. For example, \cite{entropy_segment} shows that semantic segmentation models can generalize to novel objects by following regions of concentrated attention, while \cite{grounding_entropy} demonstrates that LVLMs exhibit emergent grounding behavior when analyzing high-entropy attention heads. 
In contrast, we leverage attention entropy to suppress spurious activations for \textit{hallucination detection}. Moreover, while \cite{grounding_entropy} only accounts for the area of the visual patches, our formulation explicitly considers attention intensity to preserve vision encoder's information from multi-scale occurrences of a single object category, or small objects in the background.

\begin{figure}[t]
\centering
\caption{Performance of ADS classifier across LVLMs.}
\label{fig:ads_performance}

\resizebox{0.8\linewidth}{!}{%
\begin{tabular}{lccc}
  \toprule
  \textbf{Model} & \textbf{Prec.} & \textbf{Rec.} & \textbf{F1} \\
  \midrule
  LLaVA-1.5-7B \cite{liu2023visual}     & 0.81 & 0.74 & 0.77 \\
  InternVL-2.5-8B \cite{internvl}     & 0.79 & 0.69 & 0.73 \\
  Qwen 2.5-VL-7B  \cite{bai2025qwen03vl}     & 0.75 & 0.71 & 0.73 \\
  \bottomrule
\end{tabular}
}
\label{tab:mlp_entropy}
\end{figure}





\subsection{Cross-modal Grounding Consistency}
\label{sec:3.2_feature_similarity}

\begin{figure*}
  \centering
  \includegraphics[width=1.0\linewidth]{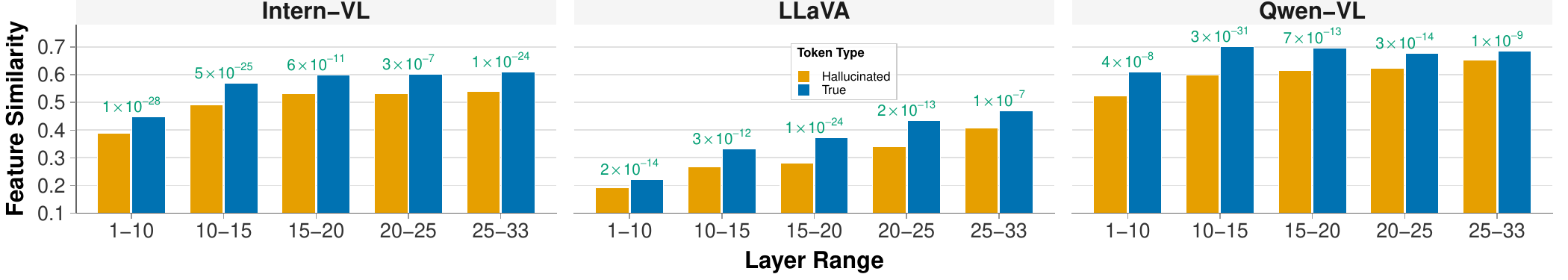}
  \caption{Layer-wise patch-level similarity scores for true vs.\ hallucinated tokens across LVLMs. Early/mid layers are most discriminative; deeper layers converge due to stronger language priors.}
  \label{fig:sim_chart}
\end{figure*}

\begin{figure}
  \centering
\includegraphics[width=1\linewidth]{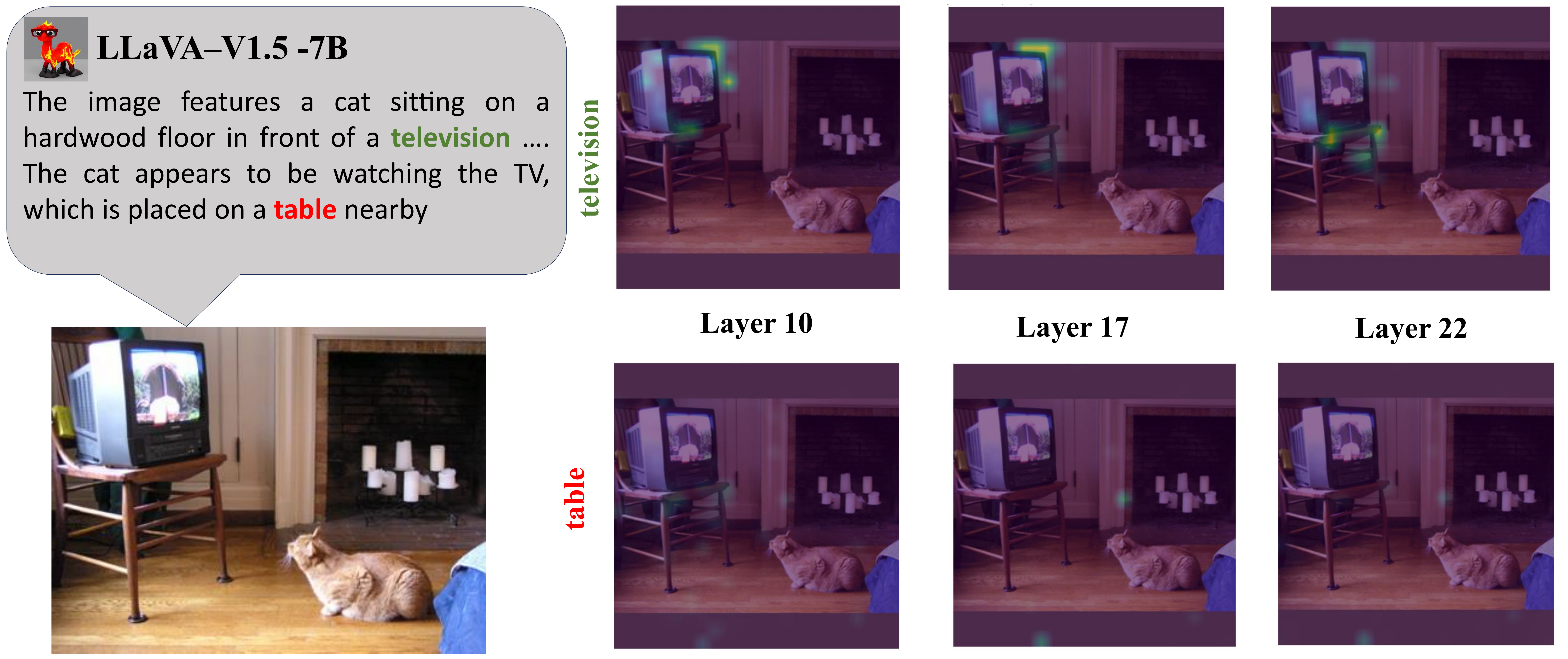}
  \caption{\textbf{Cross-modal grounding consistency heatmap} 
  for a true token (``television'', top) vs.\ a hallucinated token (``table'', bottom) across layers (10, 17, 22). True tokens display similarity clusters aligned with the object’s local structure, while hallucination tokens have low alignment with all patches across the image. }
  \label{fig:feature-similarity}
\end{figure}

\paragraph{Discovery.}
Correct tokens that correspond to visually present objects exhibit 
high peaks in patch-level similarity, while hallucinated tokens show 
low similarity over the image grid (Fig.~\ref{fig:feature-similarity}). In other words, the predicted tokens have low alignment scores with visual patches are more likely to be hallucinated.
This supports the view that the vision backbone preserves \emph{structural information}, evidenced by LVLM's true object token's features to correspond positively to their actual location. This suggests that many hallucinations originate from \emph{language priors} \cite{park2025halloc, rohrbach2018object}, thus resulting in low feature consistency with image tokens. \\
\textbf{Cross-modal Grounding Consistency} We evaluate the feature similarities between the hallucinated token and image patches; and coin this metric as \textit{Cross-modal Grounding Consistency (CGC)}. 
At layer $n$, let $h^{(n)}_t, v^{(n)}_p \in \mathbb{R}^d$ be the token and patch embeddings, and define cosine similarity
\begin{equation}
    S^{(n)}_{t,p}=\frac{\langle h^{(n)}_t, v^{(n)}_p\rangle}{\|h^{(n)}_t\|_2\|v^{(n)}_p\|_2},
\end{equation}
which reflects \emph{local structural alignment}. 
The per-token map is $\mathbf{S}^{(n)}_t=[S^{(n)}_{t,p}]_{p\in\mathcal{P}}$. 
To emphasize localized evidence, we obtain the token grounding score $C^{(n)}_t$ by aggregating the top-$k$ patches $\mathcal{T}^{(n)}_t$:
\begin{equation}
    C^{(n)}_t=\frac{1}{k}\sum_{p\in\mathcal{T}^{(n)}_t}S^{(n)}_{t,p}.
\end{equation}
\textbf{Remarks.}
Our visualized similarity maps (Fig.~\ref{fig:feature-similarity}) reveal compact peaks for grounded tokens and dispersed responses for hallucinations. We conducted layer-wise analyses across LVLMs (Fig.~\ref{fig:sim_chart}), showing strong separation in early/mid layers: true tokens average $\approx 0.37$ vs.\ hallucinations $\approx 0.29$–$0.5$, with significance confirmed by $p$-values.   The gap narrows after $\sim$layer 25, consistent with language-prior dominance at depth. The distribution of CGC is visualized in Figure \ref{fig:box_entropy}, in which hallucinated tokens displayed significantly lower CGC range than true tokens. \\
\textbf{Cross-modal Grounding Consistency as a hallucination detector.}
We built a CGC-based hallucination linear classifier to validate the effectiveness of CGC for hallucination detection, with the features being the per-layer CGC vector, similar to ADS. The results are reported in Table \ref{tab:mlp_cgc}.

\begin{figure}[tbp]
\centering
\includegraphics[width=0.65\linewidth]{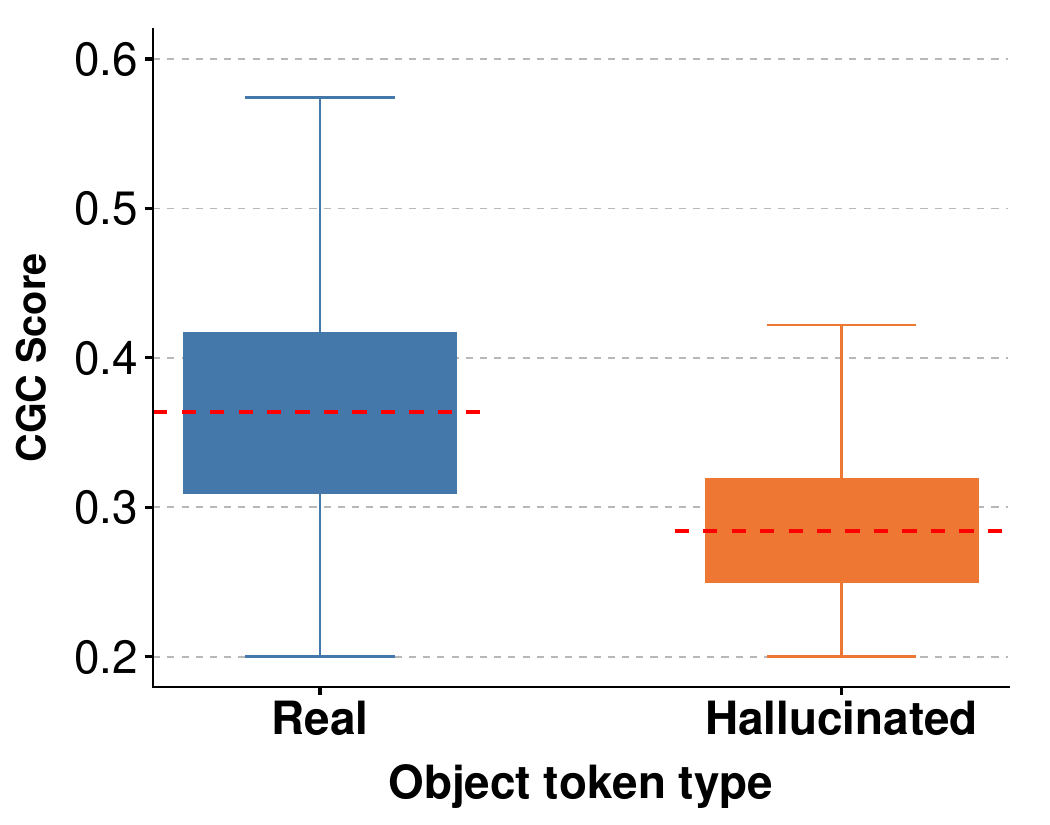}
\caption{Top--$5\%$ similarities for \emph{true} vs.\ \emph{hallucinated} tokens for LLaVA-1.5-7B.}
\label{fig:cgc_box}
\end{figure}

\begin{table}[tbp]
\centering
\caption{Performance of the CGC classifier across LVLMs.}
\resizebox{0.8\linewidth}{!}{%
\begin{tabular}{lccc}
  \toprule
  \textbf{Model} & \textbf{Prec.} & \textbf{Rec.} & \textbf{F1} \\
  \midrule
  LLaVA-1.5-7B \cite{liu2023visual}  & 0.82 & 0.76 & 0.79 \\
  InternVL-2.5-8B \cite{internvl}  & 0.80 & 0.73 & 0.76 \\
  Qwen 2.5-VL-7B \cite{bai2025qwen03vl}    & 0.83 & 0.79 & 0.81 \\
  \bottomrule
\end{tabular}
}
\label{tab:mlp_cgc}
\end{table}

\section{Token-level Hallucination Detection}
\label{sec:token_level_hallucination_detection}

\subsection{Detector's Features}
To show the effectiveness of our discovered hallucination indicators, we apply a lightweight classification model using the indicator features. 
Our approach is deliberately straightforward yet interpretable, relying on our two proposed indicators: 

\begin{enumerate}
    \item \textbf{Attention Dispersion Score} (\S\ref{sec:3.1_attention_entropy}): measures the spatial compactness of attention maps associated with an object token, after applying $k$-connected component filtering ($k=8$) to suppress noisy isolated activations. %
    Tokens grounded in the image typically show compact, low-entropy attention distributions, whereas hallucinations yield dispersed, high-entropy activations.
    \item \textbf{Cross-modal Grounding Consistency} (\S\ref{sec:3.2_feature_similarity}): quantifies the top-$k$ cosine similarity between a token embedding and visual patch embeddings, capturing structural alignment between the linguistic representation and localized visual evidence. 
    True tokens generally yield sharp similarity peaks over relevant patches, while hallucinations display weak or diffuse similarities.
\end{enumerate}

For each token $t$ generated by a LVLM, we compute both measures across all transformer layers $n \in \{1,\dots,L\}$. Denote the per-layer entropy values as
\begin{equation}
\mathbf{ADS}_t = [ADS^{(1)}_t, ADS^{(2)}_t, \dots, ADS^{(L)}_t],
\end{equation}
and the per-layer similarity scores as
\begin{equation}
\mathbf{C}_t = [C^{(1)}_t, C^{(2)}_t, \dots, C^{(L)}_t].
\end{equation}
The concatenated feature vector
\begin{equation}
\mathbf{f}_t = [\mathbf{ADS}_t \; || \; \mathbf{C}_t] \in \mathbb{R}^{2L}.
\end{equation}
thus encodes both structural attention information and local feature alignment for token $t$.  

We then train supervised classifiers to predict whether token $t$ is \emph{grounded} or \emph{hallucinated}, using $\mathbf{f}_t$ as input. We evaluate three popular Machine Learning algorithms: Extreme Gradient Boosting (XGB), Random Forest (RF) and Multi-layer Perceptron (MLP). 
Hyperparameter tuning is applied to all Machine Learning models and backbone LVLMs to ensure best performance.
\subsection{Experimental Setup}
We benchmark our detector against state-of-the-art baselines for token-level hallucination detection, namely MetaToken~\cite{fieback2024metatoken}, HalLoc~\cite{park2025halloc}, DHCP~\cite{dhcp}, ProjectAway~\cite{projectaway} and SVAR~\cite{jiang2025devils}. Implementation details are presented in the Supplementary. For our method, the concatenated feature vector $\mathbf{f}_t$ (which combines per-layer spatial entropy and feature similarities) serves as input. XGB, MLP, and Random Forest classifiers are trained on the training set, and tuned on validation data to select the best hyperparameters.
\\

\textbf{Experiments with MS COCO Image Captioning}
We first benchmark our methods against other baselines on the Image Captioning task for MS COCO images. We randomly selected 4000 images from MS COCO 2014's validation set, then split the images to have a 90-10 Train/Test split. The LVLMs are provided with the prompt \textit{"Describe this image"} to generate the description of the image. To obtain the token-level labels of true and hallucinated objects, we  use GPT-4o as the labeling model, following \cite{fieback2024metatoken}. Specifically, we provide it with the image, the MS COCO-provided human captions, and the ground truth list of objects to ensure factual grounding and reduce mislabeling. We label the token spans describing a phrase of objects, with each object instances including its prefix or accompanying objects (like \textbf{a white car} or \textbf{small birds}). Non–visual tokens (e.g., function words) are excluded from both training and inference. \\
\textbf{Experiments with POPE}
We also conduct experiments on the popular benchmark POPE \cite{pope}, with the question format being "Yes/No". Specifically, we obtain true samples by selecting LVLM's correct answers, while false responses are treated as hallucinations. We use the token "Yes" as the token for ADS, while for CGC, we extract the mentioned object token for more semantically informative hidden vectors. (e.g. the token \textit{"car"} in the question  \textit{"Is there a car in this image"}). It's worth noting that this is a heavily imbalanced classification problem, with 3339 true samples and 217 hallucinated ones, only around 6.1$\%$. Since metrics like Recall and Precision can behighly unstable and sensitive to threshold selection, we report the F1 score of the hallucination class and AUC score in this experiment after 5-fold validation for each method. 
\subsection{Experimental Results}

\vspace{-1mm}
\paragraph{Comparison with baselines for Image Captioning}

\begin{table*}[!t]
\centering
\setlength{\tabcolsep}{6pt}
\caption{Token-level hallucination detection across LVLMs for image captioning task on MS COCO dataset. \textbf{Bold} indicates best, \underline{underline} indicates second best.}
\label{tab:hallucination_results}
\resizebox{\linewidth}{!}{%
\begin{tabular}{l|cccc|cccc|cccc}
\toprule
\multirow{2}{*}{\textbf{Method}} 
& \multicolumn{4}{c|}{\textbf{LLaVA-1.5-7B}} 
& \multicolumn{4}{c|}{\textbf{Qwen2.5-VL-7B}} 
& \multicolumn{4}{c}{\textbf{InternVL2.5-8B}} \\
\cmidrule(lr){2-5} \cmidrule(lr){6-9} \cmidrule(lr){10-13}
& \textbf{PR} & \textbf{RC} & \textbf{F1} & \textbf{AUC}
& \textbf{PR} & \textbf{RC} & \textbf{F1} & \textbf{AUC}
& \textbf{PR} & \textbf{RC} & \textbf{F1} & \textbf{AUC} \\
\midrule
HalLoc~\cite{park2025halloc}        
& 0.70 & 0.73 & 0.71 & 0.76
& 0.71 & 0.73 & 0.73 & 0.80 
& 0.64 & 0.68 & 0.67 & 0.87 \\
MetaToken (LR)~\cite{fieback2024metatoken}   
& 0.61 & 0.43 & 0.51 & 0.57 
& 0.46 & 0.65 & 0.54 & 0.69 
& 0.68 & 0.62 & 0.65 & 0.60 \\
MetaToken (GB)~\cite{fieback2024metatoken}   
& 0.65 & 0.72 & 0.68 & 0.72 
& 0.70 & 0.44 & 0.54 & 0.77 
& 0.66 & \textbf{0.96} & 0.78 & 0.72 \\
SVAR~\cite{jiang2025devils}           
& 0.66 & \underline{0.89} & 0.76 & 0.75
& 0.45 & 0.78 & 0.57 & 0.68 
& 0.74 & 0.89 & 0.80 & 0.87 \\
DHCP~\cite{dhcp}
& 0.75 & 0.71 & 0.73 & 0.80
& 0.75 & 0.77 & 0.76 & 0.86
& 0.80 & 0.83 & 0.81 & 0.87 \\
ProjectAway~\cite{projectaway}
& \underline{0.80} & 0.76 & 0.79 & 0.85
& 0.78 & \underline{0.82} & \underline{0.81} & 0.89
& 0.83 & 0.81 & 0.82 & 0.89 \\
\midrule
\textbf{Ours (XGB)} 
& \textbf{0.82} & 0.81 & \textbf{0.82} & \textbf{0.88}
& \underline{0.86} & \textbf{0.83} & \textbf{0.84} & \textbf{0.94} 
& \textbf{0.88} & 0.91 & \textbf{0.90} & \textbf{0.94} \\
\textbf{Ours (MLP)} 
& 0.72 & \textbf{0.90} & \underline{0.80} & \underline{0.86}
& \textbf{0.88} & 0.76 & 0.81 & \underline{0.93} 
& \textbf{0.88} & \underline{0.92} & \textbf{0.90} & \underline{0.93} \\
\textbf{Ours (RF)}  
& 0.79 & 0.80 & 0.80 & 0.86
& 0.82 & 0.78 & 0.80 & 0.93
& \underline{0.85} & 0.91 & \underline{0.88} & 0.93 \\
\bottomrule
\end{tabular}%
}
\end{table*}

Tab.~\ref{tab:hallucination_results} compares our detector with 
existing detection methods: HalLoc \cite{park2025halloc}, MetaToken \cite{fieback2024metatoken} and SVAR \cite{jiang2025devils}
for three Large Vision Language Models, LLaVA-1.5 7B, Qwen 2.5-VL-7B, and InternVL 2.5-8B. As observed, our proposed detectors achieve state-of-the-art results across different LVLM backbones.
The CGC and ADS-based detector managed to achieve around 0.8 to 0.9 for the three representative models, demonstrating impressive hallucination detection performances.
Our method outperforms the second-best baseline consistently through all metrics and models by around 5 to 15 $\%$, most notably
by up to 19 $\%$ for LLaVA 1.5 7B  (0.85, as opposed to SVAR's 0.69). \\
\textbf{Comparison with baselines for POPE.} 
\begin{table}[t]
\centering
\begin{tabular}{l cc}
\toprule
\textbf{Method} & \textbf{F1} & \textbf{AUC} \\
\midrule
MetaToken~\cite{fieback2024metatoken} & 0.22 & 0.73 \\
HalLoc~\cite{park2025halloc}          & 0.29 & 0.71 \\
SVAR~\cite{jiang2025devils}           & 0.25 & 0.37 \\
DHCP~\cite{dhcp}                     & 0.27 & 0.69 \\
ProjectAway~\cite{projectaway}       & 0.38 & 0.70 \\
\midrule
\textbf{Ours}                        & \textbf{0.41} & \textbf{0.75} \\
\bottomrule
\end{tabular}
\caption{
Hallucination detection performance on POPE with LLaVA-1.5-7B.
Due to class imbalance, we report F1 and AUC, which provide more stable and threshold-robust evaluation.
}
\label{tab:pope}
\vspace{-5mm}
\end{table}
We report our hallucination detection results for the POPE dataset with the model LLaVA-1.5-7B with Table \ref{tab:pope}. We observe that our method obtains an F1 score of 0.41 and an AUC of 0.75, outperforming the second-best baselines \cite{projectaway} by 6 $\%$ and 2 $\%$, respectively. These results highlight the robustness of our approach across varying response formats in LVLMs. \\
\textbf{Ablation Study} 
We ablate the individual effects of ADS and CGC on our detectors' performance. We observe both of these features can return impressive accuracies of around 0.85-0.89 in AUC and 0.80-0.85 in F1 score. Combining these two powerful indicators achieves significant gains of around 3 to 6 $\%$. This confirms that the two signals are complementary: ADS captures the \textit{spatial structure} of attention over image patches — whether it is compact or dispersed — while CGC captures the \textit{semantic alignment} between the token's hidden representation and the visual content at those patches. A token can attend to a tight region yet be semantically misaligned with it (low ADS, low CGC), or attend diffusely yet happen to match relevant patches (high ADS, high CGC). Neither signal alone covers both failure modes.

\begin{table}[t]
  \centering
  \begin{tabular}{ll cc}
    \toprule
    \textbf{Model} & \textbf{Features} & \textbf{AUC} & \textbf{F1} \\
    \midrule
    \multirow{3}{*}{LLaVA-1.5}
      & ADS       & 0.86    & 0.80    \\
      & CGC       & 0.84    & 0.79    \\
      & CGC+ADS   & \textbf{0.92} & \textbf{0.84} \\
    \midrule
    \multirow{3}{*}{Qwen 2.5-VL-7B}
      & ADS       & 0.93    & 0.76    \\
      & CGC       & 0.93    & 0.80    \\
      & CGC+ADS   & \textbf{0.95} & \textbf{0.85} \\
    \midrule
    \multirow{3}{*}{InternVL 2.5-8B}
      & ADS       & 0.90    & 0.87    \\
      & CGC       & 0.93    & 0.90    \\
      & CGC+ADS   & \textbf{0.94} & \textbf{0.90} \\
    \bottomrule
  \end{tabular}
  \caption{Effect of combining CGC and ADS features on hallucination detection across three VLMs.}
  \label{tab:ablation}
\end{table}

\vspace{-3mm}
\section{Conclusion}
\label{sec:conclusion}
We present a token-level structural analysis of visual hallucinations in LVLMs and turned the resulting insights into a lightweight detector. Our proposed structural statistics: Patch-wise Attention Dispersion Score and Cross-Modal Feature Similarity, are shown to be especially diagnostic: hallucinated tokens 
exhibit scattered attention and weak alignment to any image region, whereas faithful tokens show sharp attention and localized similarity to the relevant patches. Building on these findings, we develop a detector,
achieving state-of-the-art performance 
\section*{Acknowledgement}
This research is funded by Hanoi University of Science and Technology (HUST) under grant number T2024-TĐ-002.

\clearpage

{
    \small
    \bibliographystyle{ieeenat_fullname}
    \bibliography{main}

@String(CVPR= {IEEE Conf. Comput. Vis. Pattern Recog.})

@String(NIPS= {Adv. Neural Inform. Process. Syst.})

@String(BMVC= {Brit. Mach. Vis. Conf.})

@String(AAAI = {AAAI})

@String(CVPR  = {CVPR})

@String(NIPS  = {NeurIPS})

@String(BMVC  =	{BMVC})

@inproceedings{liu2023visual,
 author = {Liu, Haotian and Li, Chunyuan and Wu, Qingyang and Lee, Yong Jae},
 booktitle = {Advances in Neural Information Processing Systems},
 editor = {A. Oh and T. Naumann and A. Globerson and K. Saenko and M. Hardt and S. Levine},
 pages = {34892--34916},
 publisher = {Curran Associates, Inc.},
 title = {Visual Instruction Tuning},
 url = {https://proceedings.neurips.cc/paper_files/paper/2023/file/6dcf277ea32ce3288914faf369fe6de0-Paper-Conference.pdf},
 volume = {36},
 year = {2023}
}

@inproceedings{
blip,
title={Instruct{BLIP}: Towards General-purpose Vision-Language Models with Instruction Tuning},
author={Wenliang Dai and Junnan Li and Dongxu Li and Anthony Tiong and Junqi Zhao and Weisheng Wang and Boyang Li and Pascale Fung and Steven Hoi},
booktitle={Thirty-seventh Conference on Neural Information Processing Systems},
year={2023},
url={https://openreview.net/forum?id=vvoWPYqZJA}
}

@ARTICLE{li2023otter,
  author={Li, Bo and Zhang, Yuanhan and Chen, Liangyu and Wang, Jinghao and Pu, Fanyi and Cahyono, Joshua Adrian and Yang, Jingkang and Li, Chunyuan and Liu, Ziwei},
  journal={IEEE Transactions on Pattern Analysis and Machine Intelligence}, 
  title={Otter: A Multi-Modal Model With In-Context Instruction Tuning}, 
  year={2025},
  volume={47},
  number={9},
  pages={7543-7557},
  doi={10.1109/TPAMI.2025.3571946}}

@inproceedings{
ye2023mplugowl,
title={m{PLUG}-Owl3: Towards Long Image-Sequence Understanding in Multi-Modal Large Language Models},
author={Jiabo Ye and Haiyang Xu and Haowei Liu and Anwen Hu and Ming Yan and Qi Qian and Ji Zhang and Fei Huang and Jingren Zhou},
booktitle={The Thirteenth International Conference on Learning Representations},
year={2025},
url={https://openreview.net/forum?id=pr37sbuhVa}
}

@misc{gemma,
      title={Gemma 3 Technical Report}, 
      author={Gemma Team and others },
      year={2025},
      eprint={2503.19786},
      archivePrefix={arXiv},
      primaryClass={cs.CL},
      url={https://arxiv.org/abs/2503.19786}, 
}

@inproceedings{
rope,
title={Multi-Object Hallucination in Vision Language Models},
author={Xuweiyi Chen and Ziqiao Ma and Xuejun Zhang and Sihan Xu and Shengyi Qian and Jianing Yang and David Fouhey and Joyce Chai},
booktitle={The Thirty-eighth Annual Conference on Neural Information Processing Systems},
year={2024},
url={https://openreview.net/forum?id=KNrwaFEi1u}
}

@article{bai2025qwen03vl,
  title={Qwen3-vl technical report},
  author={Bai, Shuai and Cai, Yuxuan and Chen, Ruizhe and Chen, Keqin and Chen, Xionghui and Cheng, Zesen and Deng, Lianghao and Ding, Wei and Gao, Chang and Ge, Chunjiang and others},
  journal={arXiv preprint arXiv:2511.21631},
  year={2025}
}

@article{internvl,
  title={Expanding performance boundaries of open-source multimodal models with model, data, and test-time scaling},
  author={Chen, Zhe and Wang, Weiyun and Cao, Yue and Liu, Yangzhou and Gao, Zhangwei and Cui, Erfei and Zhu, Jinguo and Ye, Shenglong and Tian, Hao and Liu, Zhaoyang and others},
  journal={arXiv preprint arXiv:2412.05271},
  year={2024}
}

@inproceedings{
zhu2023minigpt4,
title={Mini{GPT}-4: Enhancing Vision-Language Understanding with Advanced Large Language Models},
author={Deyao Zhu and Jun Chen and Xiaoqian Shen and Xiang Li and Mohamed Elhoseiny},
booktitle={The Twelfth International Conference on Learning Representations},
year={2024},
}

@misc{bai2023qwen,
      title={Qwen-VL: A Versatile Vision-Language Model for Understanding, Localization, Text Reading, and Beyond}, 
      author={Jinze Bai and Shuai Bai and Shusheng Yang and Shijie Wang and Sinan Tan and Peng Wang and Junyang Lin and Chang Zhou and Jingren Zhou},
      year={2023},
      eprint={2308.12966},
      archivePrefix={arXiv},
      primaryClass={cs.CV},
      url={https://arxiv.org/abs/2308.12966}, 
}

@inproceedings{pope,
  title={Evaluating object hallucination in large vision-language models},
  author={Li, Yifan and Du, Yifan and Zhou, Kun and Wang, Jinpeng and Zhao, Wayne Xin and Wen, Ji-Rong},
  booktitle={Proceedings of the 2023 conference on empirical methods in natural language processing},
  pages={292--305},
  year={2023}
}

@inproceedings{rohrbach2018object,
  title={Object hallucination in image captioning},
  author={Rohrbach, Anna and Hendricks, Lisa Anne and Burns, Kaylee and Darrell, Trevor and Saenko, Kate},
  booktitle={Proceedings of the 2018 Conference on Empirical Methods in Natural Language Processing},
  pages={4035--4045},
  year={2018}
}

@inproceedings{wu2024logical,
  title={Logical Closed Loop: Uncovering Object Hallucinations in Large Vision-Language Models},
  author={Wu, Junfei and Liu, Qiang and Wang, Ding and Zhang, Jinghao and Wu, Shu and Wang, Liang and Tan, Tieniu},
  booktitle={Findings of ACL 2024},
  year={2024}
}

@inproceedings{leng2024vcd,
  author={Leng, Sicong and Zhang, Hang and Chen, Guanzheng and Li, Xin and Lu, Shijian and Miao, Chunyan and Bing, Lidong},
  booktitle={2024 IEEE/CVF Conference on Computer Vision and Pattern Recognition (CVPR)}, 
  title={Mitigating Object Hallucinations in Large Vision-Language Models through Visual Contrastive Decoding}, 
  year={2024},
  volume={},
  number={},
  pages={13872-13882},
  doi={10.1109/CVPR52733.2024.01316}}

@inproceedings{flamingo,
  author    = {Jean-Baptiste Alayrac and Jeff Donahue and Pauline Luc and Antoine Miech and others},
title = {Flamingo: a visual language model for few-shot learning},
year = {2022},
isbn = {9781713871088},
publisher = {Curran Associates Inc.},
address = {Red Hook, NY, USA},
booktitle = {Proceedings of the 36th International Conference on Neural Information Processing Systems},
articleno = {1723},
numpages = {21},
location = {New Orleans, LA, USA},
series = {NIPS '22}
}

@inproceedings{he2025evaluating,
  title={Evaluating and Mitigating Object Hallucination in Large Vision-Language Models: Can They Still See Removed Objects?},
  author={He, Yixiao and Sun, Haifeng and Ren, Pengfei and Wang, Jingyu and Wang, Huazheng and Qi, Qi and Zhuang, Zirui and Wang, Jing},
  booktitle={Proceedings of the 2025 Conference of the North American Chapter of the Association for Computational Linguistics: Human Language Technologies (Volume 1: Long Papers)},
  address={Albuquerque, New Mexico},
  month={Apr},
  year={2025}
}

@inproceedings{park2025halloc,
  title={HalLoc: Token-level Localization of Hallucinations for Vision Language Models},
  author={Park, Eunkyu and Kim, Minyeong and Kim, Gunhee},
  booktitle={Proceedings of the Computer Vision and Pattern Recognition Conference},
  pages={29893--29903},
  year={2025}
}

@inproceedings{xiao2025detecting,
  title={Detecting and mitigating hallucination in large vision language models via fine-grained ai feedback},
  author={Xiao, Wenyi and Huang, Ziwei and Gan, Leilei and He, Wanggui and Li, Haoyuan and Yu, Zhelun and Shu, Fangxun and Jiang, Hao and Zhu, Linchao},
  booktitle={Proceedings of the AAAI Conference on Artificial Intelligence},
  volume={39},
  number={24},
  pages={25543--25551},
  year={2025}
}

@article{visualbert,
  title={VisualBERT: A Simple and Performant Baseline for Vision and Language},
  author={Liunian Harold Li and Mark Yatskar and Da Yin and Cho-Jui Hsieh and Kai-Wei Chang},
  journal={ArXiv},
  year={2019},
  volume={abs/1908.03557},
  url={https://api.semanticscholar.org/CorpusID:199528533}
}

@inproceedings{dhcp,
  title={Dhcp: Detecting hallucinations by cross-modal attention pattern in large vision-language models},
  author={Zhang, Yudong and Xie, Ruobing and Sun, Xingwu and Huang, Yiqing and Chen, Jiansheng and Kang, Zhanhui and Wang, Di and Wang, Yu},
  booktitle={Proceedings of the 33rd ACM International Conference on Multimedia},
  pages={3555--3564},
  year={2025}
}

@inproceedings{pixel_to_tokens,
  title={From pixels to tokens: Revisiting object hallucinations in large vision-language models},
  author={Shang, Yuying and Zeng, Xinyi and Zhu, Yutao and Yang, Xiao and Fang, Zhengwei and Zhang, Jingyuan and Chen, Jiawei and Liu, Zinan and Tian, Yu},
  booktitle={Proceedings of the 33rd ACM International Conference on Multimedia},
  pages={10496--10505},
  year={2025}
}

@article{woodpecker,
   title={Woodpecker: hallucination correction for multimodal large language models},
   volume={67},
   ISSN={1869-1919},
   url={http://dx.doi.org/10.1007/s11432-024-4251-x},
   DOI={10.1007/s11432-024-4251-x},
   number={12},
   journal={Science China Information Sciences},
   publisher={Springer Science and Business Media LLC},
   author={Yin, Shukang and Fu, Chaoyou and Zhao, Sirui and Xu, Tong and Wang, Hao and Sui, Dianbo and Shen, Yunhang and Li, Ke and Sun, Xing and Chen, Enhong},
   year={2024},
   month=dec }

@inproceedings{gunjal2024detecting,
author = {Gunjal, Anisha and Yin, Jihan and Bas, Erhan},
title = {Detecting and preventing hallucinations in large vision language models},
year = {2024},
isbn = {978-1-57735-887-9},
publisher = {AAAI Press},
url = {https://doi.org/10.1609/aaai.v38i16.29771},
doi = {10.1609/aaai.v38i16.29771},
booktitle = {Proceedings of the Thirty-Eighth AAAI Conference on Artificial Intelligence and Thirty-Sixth Conference on Innovative Applications of Artificial Intelligence and Fourteenth Symposium on Educational Advances in Artificial Intelligence},
articleno = {2023},
numpages = {9},
series = {AAAI'24/IAAI'24/EAAI'24}
}

@inproceedings{fieback2024metaToken,
  title={MetaToken: Detecting Hallucination in Image Descriptions by Meta Classification},
  author={Laura Fieback and Jakob Spiegelberg and Hanno Gottschalk},
  booktitle={VISIGRAPP : VISAPP},
  year={2024},
  url={https://api.semanticscholar.org/CorpusID:270094858}
}

@inproceedings{lin2014microsoft,
  title={Microsoft coco: Common objects in context},
  author={Lin, Tsung-Yi and Maire, Michael and Belongie, Serge and Hays, James and Perona, Pietro and Ramanan, Deva and Doll{\'a}r, Piotr and Zitnick, C Lawrence},
  booktitle={European conference on computer vision},
  pages={740--755},
  year={2014},
  organization={Springer}
}

@inproceedings{manakul-etal-2023-selfcheckgpt,
    title = "{S}elf{C}heck{GPT}: Zero-Resource Black-Box Hallucination Detection for Generative Large Language Models",
    author = "Manakul, Potsawee  and
      Liusie, Adian  and
      Gales, Mark",
    editor = "Bouamor, Houda  and
      Pino, Juan  and
      Bali, Kalika",
    booktitle = "Proceedings of the 2023 Conference on Empirical Methods in Natural Language Processing",
    month = dec,
    year = "2023",
    address = "Singapore",
    publisher = "Association for Computational Linguistics",
    url = "https://aclanthology.org/2023.emnlp-main.557/",
    doi = "10.18653/v1/2023.emnlp-main.557",
    pages = "9004--9017"
}

@inproceedings{jiang2025devils,
  title={Devils in middle layers of large vision-language models: Interpreting, detecting and mitigating object hallucinations via attention lens},
  author={Jiang, Zhangqi and Chen, Junkai and Zhu, Beier and Luo, Tingjin and Shen, Yankun and Yang, Xu},
  booktitle={Proceedings of the Computer Vision and Pattern Recognition Conference},
  pages={25004--25014},
  year={2025}
}

@inproceedings{entropy_segment,
author    = {Krzysztof Baron-Lis and Matthias Rottmann and Annika MÃ¼tze and Sina Honari and Pascal Fua and Mathieu Salzmann},
title     = {AttEntropy: On the Generalization Ability of Supervised Semantic Segmentation Transformers to New Objects in New Domains},
booktitle = {35th British Machine Vision Conference 2024, {BMVC} 2024, Glasgow, UK, November 25-28, 2024},
publisher = {BMVA},
year      = {2024},
url       = {https://papers.bmvc2024.org/0215.pdf}
}

@inproceedings{grounding_entropy,
  title={Your large vision-language model only needs a few attention heads for visual grounding},
  author={Kang, Seil and Kim, Jinyeong and Kim, Junhyeok and Hwang, Seong Jae},
  booktitle={Proceedings of the Computer Vision and Pattern Recognition Conference},
  pages={9339--9350},
  year={2025}
}

@inproceedings{
projectaway,
title={Interpreting and Editing Vision-Language Representations to Mitigate Hallucinations},
author={Nicholas Jiang and Anish Kachinthaya and Suzanne Petryk and Yossi Gandelsman},
booktitle={The Thirteenth International Conference on Learning Representations},
year={2025},
url={https://openreview.net/forum?id=94kQgWXojH}
}
}

\clearpage
\setcounter{page}{1}
\maketitlesupplementary

\section{Experimental configurations}
For hallucination detection, our configurations are as follows:
\begin{itemize}
    \item \textbf{MetaToken}: For our greedy decoding setup, the probability difference term in Eq.~11 of the original paper is always zero; we instead use token probability directly. We implement two binary classifiers, Logistic Regression (LR) with \texttt{lbfgs} solver and Gradient Boosting (GB) with 100 estimators.
    \item \textbf{HalLoc}: We use a pretrained CLIP-ViT-B/32 encoder with a linear projection to match VisualBERT input dimensions. A single classification head is used, as we focus exclusively on object hallucination. Training follows the original optimizer setup (AdamW, $\beta=(0.9,0.999)$, weight decay $1.0 \times 10^{-2}$, learning rate $1.0 \times 10^{-6}$) with batch size 16 on an RTX 3090.
    \item \textbf{SVAR}: We train a one-hidden-layer MLP (hidden dim 248, learning rate $0.001$) for 50 epochs as a result of the hyperparameter search strategy.
\end{itemize}

\subsection{LVLM Inference Parameters}

All models use temperature $=0.1$ and top-$p = 20$, top-$p = 50$ for decoding.

\subsection{Classifier Hyperparameters}

We optimize XGB, RF and MLP classifiers via grid search.

\begin{table}[h]
\centering
\begin{tabular}{l l l}
\toprule
\textbf{Model} & \textbf{Hyper‑parameter} & \textbf{Values} \\
\midrule
\multirow{3}{*}{XGB} 
 & depth        & \{4,6,8\} \\
 & learning rates           & \{0.1,0.05\} \\
 & no. estimators   & \{100,200,500\} \\[0.2em]
\midrule
\multirow{3}{*}{RF}
 & depth        & \{None,10,20\} \\
 & trees        & \{200,400,600\} \\
\midrule
\multirow{3}{*}{MLP}
 & hidden size      & \{64,128,256\} \\
 & learning rate           & \{0.01,0.001\} \\
 & optimizer    & \{\text{Adam},\text{SGD}\} \\
\bottomrule
\end{tabular}
\caption{Hyper‑parameter ranges for each classifier.}
\label{tab:hp_summ_no_f1}
\end{table}

The best hyper-parameter configurations were found to be: for XGB, a depth of 6, learning rate of 0.05, and 500 estimators; for RF, a maximum depth of 10 with 400 trees; and for MLP, a hidden layer size of 128, a learning rate of 0.001, and the Adam optimizer.

\section{Dataset curation}
In order to construct a hallucination detection dataset, instead of following previous papers to use CHAIR~\cite{rohrbach2018object}, we used GPT-4o API to extract hallucinated words. This is because the CHAIR toolkit often misses or return excessive words in case of ambiguity. The prompt structure we used is as follows \ref{box:hallucination_prompt}:





\begin{figure}[t]
    \centering
    \includegraphics[width=0.99\linewidth]{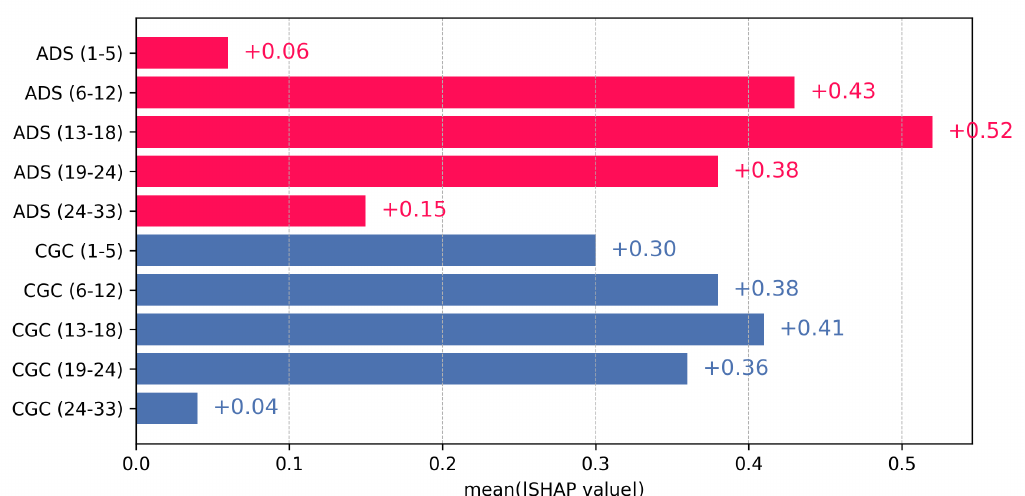}
    \caption{Mean absolute SHAP values for ADS and CGC across layers.}
    \label{fig:ads_cgc_shap}
\end{figure}

It is possible that an object word can consist of more than 1 token, therefore we calculate the token-level attribute by indexing the object word, and obtain a `prefix` which is the sentence prior to that word. Then the VLMs will forward once based on the `prefix`, yielding the internal attributes of the object token.

\section{Ablation Studies}
\label{sec:ablation}
We keep the classifier fixed (LLaVA features + MLP) and vary which layer combinations are included.  
The results for our two features are shown in \cref{tab:layer_ablate}.  
We observe that, typically the middle layers (around 12-24) are the layers with richest semantic features and best alignment between text and images, yielding the best results. Meanwhile, the last layers are usually used to refine the language predictions of the Language Encoder, rendering visual information to be weaker and leading to poor classification results.

\begin{table}[h]
\centering
\begin{tabular}{lcccc}
\toprule
\multirow{2}{*}{Layer Set} & \multicolumn{2}{c}{ADS} & \multicolumn{2}{c}{CGC} \\
\cmidrule(r){2-3} \cmidrule(l){4-5}
 & AUC & F1 & AUC & F1 \\
\midrule
Layers 6–12   & 72.38 & 67.88 & 76.51 & 73.26 \\
Layers 13–18  & 69.17 & 62.91 & 73.34 & 70.55 \\
Layers 19–24  & 68.31 & 63.74 & 68.79 & 63.44 \\
Layers 25–33  & 64.58 & 62.88 & 75.44 & 74.34 \\
All layers    & \textbf{85.84} & \textbf{80.36} & \textbf{84.41} & \textbf{78.90} \\
\bottomrule
\end{tabular}
\caption{Layer-wise ablation of ADS and CGC features.}
\label{tab:layer_ablate}
\end{table}




\section{Feature Importance}

We compute the mean SHAP values of layer-wise ADS and CGC features, for LLaVA-1.5 + MLP classifier. The result is displayed with Fig \ref{fig:ads_cgc_shap}. We observe the findings is consistent with \ref{sec:ablation}, where features from the middle layers are the most influential for cross-modal attention and alignment, meanwhile for the final layers these information are no longer encoded, leading to poor performance and low predictive importance.


The overthinking-based CGC and ADS features dominate the contribution, confirming that they encode the most discriminative signal.

\section{Combining Our Features With SVAR}
\begin{table}[h]
\centering
\begin{tabular}{llcc}
\toprule
Model & Features & AUC & F1 \\
\midrule
\multirow{4}{*}{LLaVA-1.5} 
  & SVAR          & 85.12 & 69.35 \\
  & +CGC          & 87.72 & 73.52 \\
  & +ADS          & 89.04 & 74.03 \\
  & +CGC+ADS      & \textbf{89.41} & \textbf{74.07} \\
\midrule
\multirow{4}{*}{Qwen 2.5-VL-8B } 
  & SVAR          & 87.85 & 76.38 \\
  & +CGC          & 86.79 & 76.24 \\
  & +ADS          & 89.11 & 79.12 \\
  & +CGC+ADS      & \textbf{90.28} & \textbf{81.09} \\
\midrule
\multirow{4}{*}{InternVL-2.5-8B} 
  & SVAR          & 86.21 & 76.31 \\
  & +CGC          & 86.39 & 79.43 \\
  & +ADS          & 88.02 & 81.03 \\
  & +CGC+ADS      & \textbf{89.54} & \textbf{81.56} \\
\bottomrule
\end{tabular}
\caption{Effect of adding CGC and ADS features to hallucination detectors across three VLMs.}
\label{tab:model_cgc_ads}
\end{table}

We combine our metrics (ADS + CGC) with the prior attention-based detection method, SVAR.   
Specifically, we add all of our proposed features alonside with SVAR's original features, and train a MLP as a classifier.
Results are shown in  \cref{tab:model_cgc_ads}. We observe that adding our crafted features help yield upto 5 $\%$ in performance gain, proving the effectiveness of our method, and also demonstrate how our proposed metrics can be applied along with other classifiers.



\section{Ablation Studies}
We report the sensitivity of our detector to two important threshold parameter choices: Top-$x$\% (Equation 1, page 5), which determines how many attention patches we recognize as object patches with 8-connected components for ADS, and Top-$k$ for CGC (Equation 7), which are the percentage of high semantic similarity areas that we aggregate. The experiments are conducted on LLaVA-1.5-7B.As observed in Table \ref{tab:threshold_sensitivity}, we see that for the value around 5-30 $\%$, the detector manages to achieve a stable performance of around 80-85 \% in AUC and around 75-80 \% in F1 score. The performance starts to degrade after 15\%, since selecting too many patches can lead to attention noises and sinks being selected, worsening the discriminative signals. The same phenomenon is observed for CGC, where the accuracy is stable from around 1-10\%, but deteriorates heavily to a F1 of 72.96 when we select 20 \% of the most semantically similar patches.   

\begin{table}[t]
  \centering
  \begin{tabular}{ccc|ccc}
    \toprule
    \multicolumn{3}{c|}{\textbf{ADS}} & \multicolumn{3}{c}{\textbf{CGC}} \\
    \cmidrule(lr){1-3} \cmidrule(lr){4-6}
    \textbf{Top-$x$\%} & \textbf{AUC} & \textbf{F1} & \textbf{Top-$k$\%} & \textbf{AUC} & \textbf{F1} \\
    \midrule
    5\%   & 81.65 & 76.54 & 1\%   & 81.36 & 75.31 \\
    \textbf{10\%}  & \textbf{83.86} & \textbf{76.83} & 3\%   & 82.55 & 76.54 \\
    15\%  & 82.87 & 77.65 & \textbf{5\%}   & \textbf{83.86} & \textbf{76.83} \\
    20\%  & 81.65 & 76.07 & 10\%  & 82.57 & 77.78 \\
    30\%  & 79.83 & 74.70 & 20\%  & 81.82 & 72.96 \\
    \bottomrule
  \end{tabular}
  \caption{Sensitivity of ADS foreground threshold and CGC top-$k$\% patch selection on LLaVA-1.5-7B. Each side varies one hyperparameter while fixing the other at its default. Bold indicates default.}
  \label{tab:threshold_sensitivity}
\end{table}

\begin{promptbox}{LLM-based Hallucination Detector}
\label{box:hallucination_prompt}
\textbf{System Prompt:} \\
You are a precise hallucination detector. Follow the instructions exactly and output ONLY a JSON list. \\
\vspace{0.5em}
\textbf{User Prompt:} \\
You are given:  \\
- A list of ground truth object classes (from COCO).  \\
- A detailed description of an image.  \\
- Several captions of the same image.  \\
Your task:  \\
Find all object classes that are mentioned in the description, but are NOT mentioned in any of the captions, and are NOT present in the ground truth list.  \\
Output the result as a list as in the examples. Do NOT add any extra text or provide any explanations.  \\
\textbf{Examples:}  \\
\textit{Objects:} ["bowl", "broccoli", "carrot"]  \\
\textit{Description:} There are two bowls of food, one containing a mix of vegetables, such as broccoli and carrots, and the other containing meat.  \\
\textit{Captions:}  \\
- A bowl with broccoli and carrots.  \\
$\rightarrow$ Output: ["meat"]  \\
\vspace{0.5em}
\textit{Objects:} ["bowl", "broccoli"]  \\
\textit{Description:} A bowl full of broccoli.  \\
\textit{Captions:}  \\
- A bowl of green vegetables.  \\
$\rightarrow$ Output: []  \\

\vspace{0.5em}

\textbf{Now answer:}  \\
\textit{Objects:} \{objects\}  \\
\textit{Description:} \{description\}  \\
\textit{Captions:}  \\
\{captions\_formatted\}  \\
$\rightarrow$ Output:

\end{promptbox}

\end{document}